\documentclass{ecai} 

\usepackage{times}
\usepackage{soul}
\usepackage{url}
\usepackage[hidelinks]{hyperref}
\usepackage[utf8]{inputenc}
\usepackage[small]{caption}
\usepackage{graphicx}
\usepackage{amsmath}
\usepackage{amsthm}
\usepackage{booktabs}
\usepackage{algorithm}
\usepackage{algorithmic}
\usepackage[switch]{lineno}
\usepackage{comment}
\usepackage{amssymb}
\usepackage{amsfonts}
\usepackage{multirow}
\usepackage{enumitem}
\usepackage{subcaption}
\usepackage{float}

\newcommand{\BibTeX}{B\kern-.05em{\sc i\kern-.025em b}\kern-.08em\TeX}

\begin{document}


\begin{frontmatter}



\title{Towards Human-Like Grading: A Unified LLM-Enhanced Framework for Subjective Question Evaluation}

\author[A]{\fnms{Fanwei}~\snm{Zhu}\footnote{Co-first authors with equal contribution.}}
\author[B]{\fnms{Jiaxuan}~\snm{He}\footnotemark}
\author[C]{\fnms{Xiaoxiao}~\snm{Chen}\thanks{Corresponding Author. Email: 814441073@qq.com.}} 
\author[B]{\fnms{Zulong}~\snm{Chen}}
\author[D]{\fnms{Quan}~\snm{Lu}} 
\author[A]{\fnms{Chenrui}~\snm{Mei}} 

\address[A]{Hangzhou City University}\address[B]{Alibaba Group}\address[C]{Zhejiang Hospital}\address[D]{Mashang Consumer Finance Co}

\begin{abstract}
Automatic grading of subjective questions remains a significant challenge in examination assessment due to the diversity in question formats and the open-ended nature of student responses. Existing works primarily focus on a specific type of subjective question and lack the generality to support comprehensive exams that contain diverse question types. In this paper, we propose a unified Large Language Model (LLM)-enhanced auto-grading framework that provides human-like evaluation for all types of subjective questions across various domains. Our framework integrates four complementary modules to holistically evaluate student answers. In addition to a basic text matching module that provides a foundational assessment of content similarity, we leverage the powerful reasoning and generative capabilities of LLMs to: (1) compare \textit{key knowledge points} extracted from both student and reference answers, (2) generate a \textit{pseudo-question} from the student answer to assess its relevance to the original question, and (3) simulate human evaluation by identifying content-related and 
non-content strengths and weaknesses.
    Extensive experiments on both general-purpose and domain-specific datasets show that our framework consistently outperforms traditional and LLM-based baselines across multiple grading metrics. Moreover, the proposed system has been successfully deployed in real-world training and certification exams at a major e-commerce enterprise. 
    
\end{abstract}
\end{frontmatter}
\newcommand{\eg}{\textit{e.g.}}
\newcommand{\xeg}{\textit{E.g.}}
\newcommand{\ie}{\textit{i.e.}}
\newcommand{\xie}{\textit{I.e.}}
\newcommand{\etc}{\textit{etc.}}
\newcommand{\etal}{\textit{et al.}}
\newcommand{\wrt}{w.r.t. }
\newcommand{\mb}[1]{\mathbf{#1}}
\newcommand{\mc}[1]{\mathcal{#1}}
\newcommand{\figref}[1]{Fig.~\ref{#1}}
\newcommand{\tableref}[1]{Table~\ref{#1}}
\newcommand{\sectref}[1]{Sect.~\ref{#1}}
\newcommand{\B}[2]{\mathcal{B}_{#1}^{#2}} 
\newcommand{\xs}[2]{\mathcal{S}_{#1}^{#2}} 
\newcommand{\s}[2]{s_{#1}^{#2}} 
\newcommand{\x}[1]{\mathbf{x}_{#1}} 
\newcommand{\uv}[2]{\mathbf{v}_{#1}^{#2}} 
\newcommand{\iv}[1]{\mathbf{v}_{#1}}  
\newcommand{\fs}{g}   
\newcommand{\samp}{\mathbb{Y}}
\newcommand{\stitle}[1]{\vspace{2mm} \noindent {\bf #1}}
\newcommand{\sstitle}[1]{\vspace{1.5mm} \noindent {\emph{#1}}}
\newcommand{\sa}{SASNet}
\newcommand{\vp}[2]{\mathbf{c}_{#1}^{#2}} 
\newcommand{\pop}{\beta} 
\newcommand{\com}{\alpha} 
\newcommand{\ours}{LLMAG}

\renewcommand{\vec}{\bold}
\newcommand{\priceV}{\vec{z}}
\newcommand{\qualityV}{\vec{q}}
\newcommand{\incomV}{\vec{i}}
\newcommand{\excomV}{\vec{e}}
\newcommand{\lpopV}{\vec{l}}
\newcommand{\spopV}{\vec{u}}
\newcommand{\comV}{\vec{c}}
\newcommand{\popV}{\vec{p}}

\section{Introduction}
\label{sec-intro}
Automatic examination assessment has been widely adopted in educational institutions and online testing platforms to reduce grading effort, accelerate feedback processes, and ensure unbiased evaluation~\cite{babitha2022trends}. Subjective questions, such as short-answer questions, problem-solving tasks, and case analyses, constitute a central component of comprehensive examinations. However, grading subjective questions is inherently more complex and labor-intensive than evaluating objective ones due to their diverse formats, ambiguous semantics, and open-ended nature.

Existing works on automatic subjective question evaluation primarily focus on specific question types. For example, automatic essay scoring (AES) ~\cite{li2024automated} evaluate long-form essays using rubric-based criteria while automatic short answer grading (ASAG)~\cite{zhang2022automatic} emphasizes factual alignment with reference answers. These methods lack the generality to grade exams containing diverse question types. Technically, even recent LLM-based efforts~\cite{gu2024survey} often reduce grading to simplistic similarity comparison or prompted scoring, missing the question-adaptable, context-aware judgment of human graders- such as rewarding creativity for essays or precision for technical problems.
Therefore, there remains a critical gap in building a unified grading framework capable of \textit{handling the diversity and complexity of real-world subjective questions with human-like judgment}.

\stitle{Challenges.} Designing such a unified, comprehensive auto-grading system for all types of subjective questions involves multiple core challenges:
\begin{itemize}[leftmargin=*]
\item \emph{Redundancy and ambiguity.} Student answers often contain irrelevant or repetitive information, making it difficult to identify the most relevant points for accurate grading. Moreover, ambiguities in phrasing or interpretation further complicate the assessment of subjective question. 
\item \emph{Holistic, multi-faceted assessment.} Unlike objective questions that have clear, determined answers, subjective responses require comprehensive assessment of both content aspects (\eg, factual accuracy, completeness) and non-content aspects (\eg, logical coherence, clarity, and structure), resembling the cognitive processes of human graders. 
\item \emph{Weak answer alignment.} Students' answers can significantly deviate from the reference materials in terms of length, wording, and perspective, yet remain relevant and insightful. Direct answer-to-answer alignment is therefore insufficient, as responses that appear quite different to the reference answers may still be relevant to the question in many cases.
\end{itemize}

\stitle{Proposal.} To address these challenges, we propose a \emph{unified LLM-enhanced framework} for automatic grading of diverse subjective questions. Unlike direct LLM prompting approaches, our framework integrates multiple complementary modules that leverage LLMs' advanced language understanding and generation capabilities to enhance grading accuracy and robustness.

\emph{First}, LLMs excel at distilling the most pertinent content while filtering redundant information. This capability motivates us to develop a \emph{Key Points Matching Module} (KPM) that identifies the essential knowledge points from the student and reference answers for a \emph{knowledge-level alignment}, improving grading accuracy despite answer redundancy and ambiguity (Challenge 1).

\emph{Second}, trained on vast amounts of text data, LLMs possess robust capabilities in comprehending and evaluating nuanced language expressions. We leverage this through a \textit{LLM-based General Evaluation Module} (LGE) that assesses answers across multiple dimensions (\eg, semantic relevance, logical coherence, and clarity of expression) to simulate human graders' comprehensive judgment (Challenge 2).

\emph{Third}, to address weak answer alignment, we introduce a novel reverse-matching strategy through a \textit{Pseudo-Question Matching Module} (PQM). This module capitalizes on LLMs' generative capabilities to create pseudo-questions from students' answers, then evaluate their semantic alignment with the original question. Such \textit{question-to-question} comparison overcomes the limitations of direct answer matching when student responses diverge in structure or phrasing yet remain semantically valid (Challenge 3).
 
\emph{Additionally}, we incorporate a \textit{Textual Similarity Matching Module} (TSM) that provides direct lexical comparisons between student and reference answers to ensure no critical textual details are overlooked during the semantic evaluation. Finally, a \textit{Deep Fusion Layer} attentively integrates insights from all four modules, capturing cross-module dependencies and interactions for final scoring.

\stitle{Evaluation.} Due to the absence of public benchmarks for mixed-type subjective question assessment, we construct two novel datasets: one from educational examinations spanning four academic disciplines, and another from enterprise certification tests featuring domain-specific questions. To ensure comprehensive evaluation, we conduct extensive experiments on these datasets alongside a widely-used AES benchmark. Results show that our method consistently outperforms strong baselines across diverse question types, domains, and evaluation metrics.

In summary, our main contributions are as follows.
\begin{itemize}
\item We identify key limitations of existing approaches and propose a unified, LLM-enhanced framework that simulates {human-like grading} across all types of subjective questions.
\item We develop four complementary modules that collectively capture both content and expression quality, combining semantic alignment and textual similarity to support comprehensive evaluation.
\item We build and release two evaluation datasets covering diverse domains and question types, enabling rigorous benchmarking.
\item We validate the effectiveness of our approach through extensive experiments and real-world deployment. Our code and datasets are released to support future research~\cite{code}.
\end{itemize}


\section{Related work}
\label{sec-related}



Existing works on automatic subjective question grading can be categorized into two types: Automatic Short Answer Grading (ASAG) which focuses on short, concise responses to questions with a pre-defined reference answer~\cite{zhang2022automatic}  and Automatic Essay Scoring (AES) that typically deals with longer, structured essays written on a given topic~\cite{ramesh2022automated}. Although ASAG and AES differs in terms of the reliance on correct answers, the techniques in both generally evolves from traditional rule-based models to the state-of-the-art learning-based models.





\stitle{Automatic Short Answer Grading}. The primary goal of ASAG is to score the responses accurately against reference answers and its techniques emphasize on the semantic similarity with the reference answer. 
Traditional works mainly utilized handcrafted features for similarity evaluation.
For instance, He~\etal~\cite{he2009automatic} proposed to integrate LSA and n-gram co-occurrence for improving the accuracy of automatic summary assessment. 
Das \etal~\cite{das2022automatic} considered the string similarity, semantic similarity, and keyword similarity as the criterion and proposed a weighted multi-criteria-decision-making
(MCDM) approach to integrate all these similarities for subjective assessment.
Recently, deep learning approaches and LLMs have been widely explored for ASAG. For instance, 
Zhu \etal~\cite{zhu2022automatic} developed a BERT-based neural network that integrates dynamic text encoding, a semantic refinement layer, and a novel triple-hot loss strategy to enhance semantic understanding and scoring precision. Yoon \etal~\cite{yoon2023short} used LLM to identify the key phrases in student answers and compare them with those of reference answers for scoring. Schneider \etal~\cite{schneider2023towards} used GPT 3.5 to assess instructor answers, student answers, and their similarity, reporting observed issues with the LLM grading compared to human assessment.

\stitle{Automatic Essay Scoring.}
AES aims to provide a holistic evaluation in multi-paragraph essays, focusing on writing quality, coherence, and content relevance.
Traditional AES systems heavily depended on rule-based heuristics or feature-based similarity metrics~\cite{attali2006automated,bin2008automated}. The availability of annotated corpora later prompted a shift toward learning-based AES approaches. 
Transformer-based models, such as R2BERT \cite{yang2020enhancing}, multi-scale BERT \cite{wang2022use}, further advanced this area by leveraging pre-trained linguistic and commonsense knowledge.
LLMs have introduced new possibilities for AES through prompting techniques, allowing models to score essays with minimal supervision. For example,
Lee \etal~\cite{lee2024prompting} used LLMs in a zero-shot setting, where no manually scored essays were provided for training. Mansour \etal~\cite{mansour2024can} and Xiao \etal~\cite{xiao2024automation} investigated the effectiveness of LLMs for scoring in a few-shot setting, where a small set of labeled examples were included in the prompt to guide the scoring process. Song \etal ~\cite{song2024automated} explored the use of open-source LLMs for AES and automated essay revising (AER), highlighting LLMs as efficient, cost-effective, and privacy-friendly solutions for AES and AER tasks.

\stitle{Comparison to our work.} Our work differs from existing studies in two aspects: \emph{First}, while existing works focus on specific type of question, we propose a uniform auto-grading framework capable of handling all types of subjective questions. This include both questions with targeted or factual answers and essay-writing question with open-ended responses.
\emph{Second}, existing methods on the use of LLMs for auto-scoring are still in the early stages, showing poor alignment with human graders. Instead of using LLMs as a tool for scoring, we develop an LLM-enhanced method that fully explores the ability of LLMs in different evaluation aspects to simulate the human graders' consideration and thus enhance the accuracy of auto-grading.


\section{Preliminaries and Problem}
\label{sec:problem}


The application scenarios of subjective question auto-grading span across various domains such as education, corporate training, and specialized assessment. Subjective questions in comprehensive exams are designed to assess not only factual knowledge but also analytical skills, critical thinking, problem-solving, and the ability to apply knowledge. Automating the grading of such exams requires consideration of all possible types of subjective questions, including short answer questions (\eg, definition, noun explanation), writing-based questions (\eg, essay writing), and scenario questions (\eg, case study). 
Based on the availability of correct answers, we categorize these question into two broad types: \textit{factual-type questions} with fixed reference answers (\eg, short answer question) and \textit{open-ended questions} with only scoring rubrics (\eg, essay-type question).   
In this paper, we aims to provide a unified solution for auto-grading across all types of subjective questions.

\stitle{Subjective question auto-grading.} Given a subjective question $Q$, a student answer $A$ to the question, and the reference materials $R$, which may include a reference answer, scoring points, or rubrics, the task is to construct a scoring model $\mathcal{F}$ that generates the predicted score $\hat{y}$ of $A$ based on $Q$ and $R$:
\begin{align}
     \mathcal{F}(A,Q,R) = \hat{y}
     \vspace{-2mm}
\end{align}
with the objective of minimizing the difference between $\hat{y}$ and the true score $y$ provided by the human grader.

\section{LLM-powered Auto-grading Approach}
\label{sec:method}
\subsection{Overall framework}

\begin{figure}[h]
    \centering
    \includegraphics[width=\columnwidth]{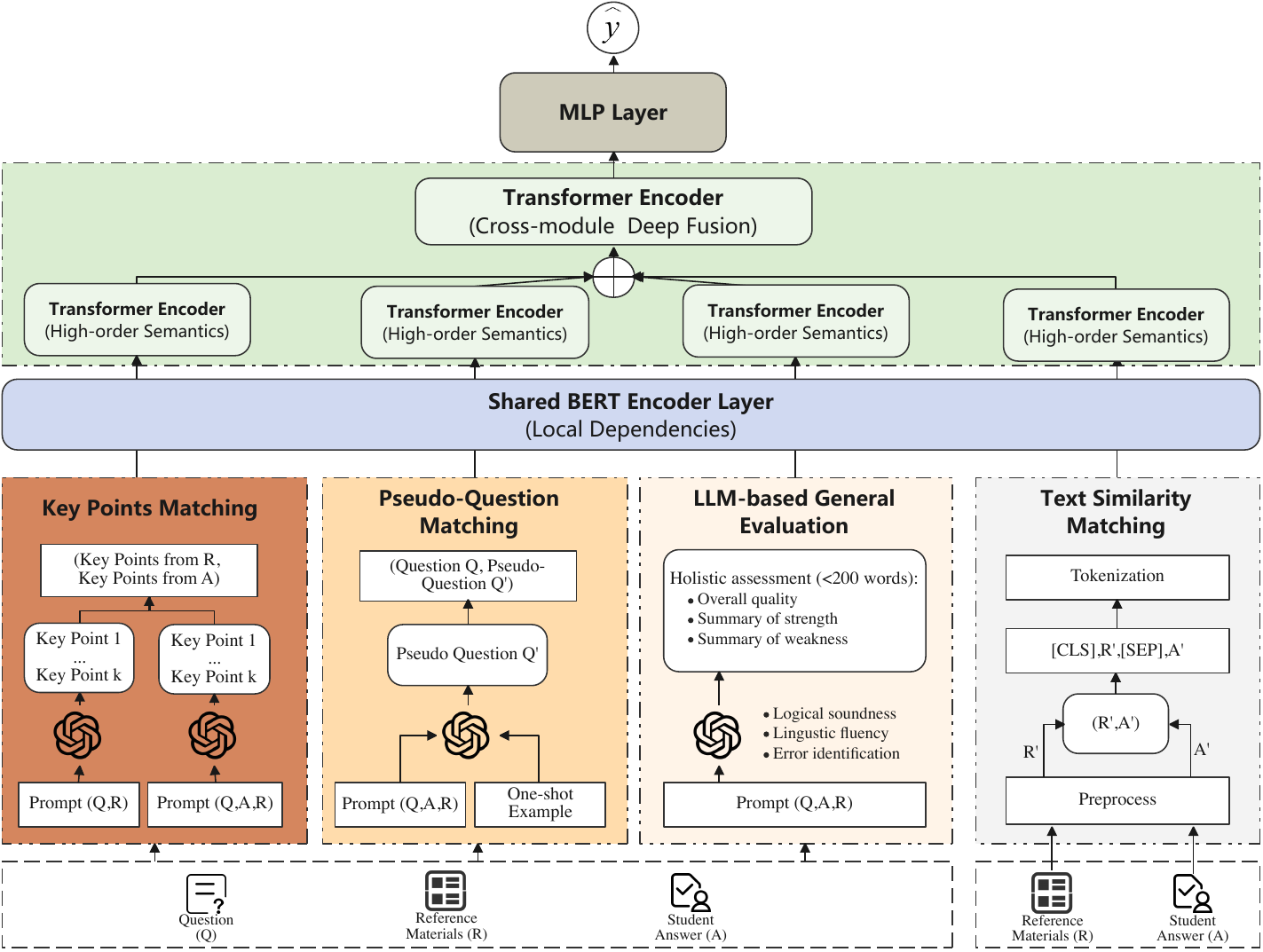}
    \vspace{2mm}
    \caption{The architecture of LLM-powered auto-grading model.}
    \label{fig:archi}
\end{figure}

The overall architecture of the proposed LLM-enhanced auto-grading framework is illustrated in Fig.~\ref{fig:archi}. It consists of four key branches: (1) \textit{Key Points Matching Module} that focuses on reducing redundancy and noise in the student answer by extracting key knowledge points from the student and reference answers with LLM; (2) \textit{Pseudo-Question Generation and Matching Module} that utilizes LLM to generate pseudo-questions based on the student's answer, which are then compared with the original question to evaluate how well the student answer written in various formats aligns with the intent of the question; (3) \textit{LLM-based General Evaluation Module} that leverages the understanding and reasoning capability of LLM to evaluate the alignment of the response with the reference answer  across multiple dimensions; (4) \textit{Textual Similarity Matching Module} that performs direct text matching between the student answer and reference answer using similarity matching technique, ensuring that no critical details are missing during feature extraction by the LLM. 
The output of each module is encoded and then integrated in a \textit{Deep Fusion Layer} to produce the final score.

\subsection{Key Points Matching}
In student answers of subjective questions, the same idea or knowledge may be expressed in diverse ways. Some may include unnecessary details, unclear phrasing, or irrelevant information that distract from the core concept. Moreover, for essay-writing type questions, reference answers are usually presented as scoring guidelines rather than standard answers (\eg, 2 points for optimism, 1 point for perseverance), which makes direct comparison with reference answer even more infeasible.

To address these issues, we design a key points matching module that leverages LLM to extract the key points from student answers and reference answers, removing redundancy and noise. The extracted key points are combined as a fused representation, which is passed into a BERT model to learn the contextual embedding for semantic matching.

\sstitle{Key points extraction.} Given the question $Q$, the student answer $A$ and the reference answer $R$ as input, the LLM is prompted to analyze and extract essential knowledge points that match the intent of the question. In particular, for the student answer $A$, the LLM identifies key points that summarize the main ideas, taking into account the context provided by both the question and the reference answer. For the reference answer $R$, the LLM generates a comparable set of key points in alignment with the scoring criteria. If the reference answer only consists of scoring guidelines but no concrete answers, the guidelines are condensed and returned as key points.

This step ensures that the extracted knowledge points from both answers are concise, representative, and semantically relevant to the question. We formalize this as:
\begin{align}
    K_A = ExtractKeyPoints (Q, A, R) \\
    K_R = ExtractKeyPoints (Q, R)
    \vspace{-2mm}
\end{align}
where $K_A$ and $K_R$ are the key points extracted from $A$ and $R$ respectively, typically in the form of 2-3 phrases, each within 25 characters.

\sstitle{Key points pairing and encoding.} To capture the semantic correlation between student and reference answers, the extracted knowledge points are organized into a pair $<K_A, K_R>$, which is tokenized (\ie, $[CLS]K_A [SEP]K_R[SEP]$) and passed into a BERT model to obtain the contextual embeddings $\mathbf{H_B}$, formalized as:
\begin{align}
    \mathbf{B} = \mathit{BERT}(<K_A, K_R>)
    \vspace{-2mm}
\end{align}
Here, $\mathbf{H_B}\in \mathbb{R}^{L\times d}$ where $L$ is the length of tokenized sequence and $d$ is the hidden size.
While BERT captures the local semantics in the token-level embeddings, we augment it with a Transformer layer to better model high-order interactions between $K_R$ and $K_A$: 
\begin{align}
\vspace{-2mm}
    \mathbf{H}_K= \mathit{Transformer}(\mathbf{B})
    \vspace{-1mm}
\end{align}

\subsection{Pseudo-Question Generation and Matching}
While knowledge points matching module focuses on the semantic similarity between student answer and reference answer to tackle the redundancy issue, we also notice that assessing answers alone cannot verify true relevance to the question in many cases. For example, the responses to open-ended questions could be highly diverse with varying length and different opinions, direct answers alignment may cause \textit{false negatives} where student answers may appear differently to reference answer but are actually related to the question.

To reduce these risks, we introduce the Pseudo-Question Generation and Matching module, leveraging the capabilities of LLM to generate a pseudo-question based on the student's answer, capturing the intent of the response. We then assess the semantic alignment between pseudo-question and the original question, which indirectly measures whether the student answer is inherently related to the question. Such reversed matching allows a dual check of answer similarity and query relevance and improves grading accuracy. 

\sstitle{Pseudo-question generation with one-shot prompt.}
Considering that general-purpose LLMs may face limitations when generating domain-specific questions in specialized exams, we use one-shot prompting that provides LLM with an example of question-answer pair from the same domain to guide the generation of pseudo-questions. Specifically, for a given student answer $A_i$, we first conduct a similarity search within the question pool $\mathbb{Q}$, which consists of all <question, reference answer> pairs from the dataset. We encode both student and reference answers using BERT, compute their cosine similarity, and retrieve the most relevant question-answer pair from $\mathbb{Q}$ as an example, denoted as ${e}_i=(Q_s, A_s)$. 

Next, the one-shot prompt is constructed, including the retrieved question-answer pair $e_i$ and the student answer $A_i$, to guide the LLM generate relevant and domain-aware pseduo-question $Q'_i$ for $A_i$:
\begin{align}
\vspace{-2mm}
    Q'_i = LLM ({e}_i,A_i,Q_i)
\end{align}
Here, the LLM is specifically prompted to generate high-quality, semantically relevant pseudo-questions while avoiding direct retrieval of the original question $Q_i$. 

\sstitle{Semantic alignment of question pairs.} 
The generated question and original question is formed into a pair $<Q'_i, Q_i>$ for similarity alignment. This step is similar as in the key points matching module: the question pair is passed through the shared BERT layer followed by a Transformer layer to model both local and global semantic relevance between the two questions, formalized as:
\begin{align}
\vspace{-2mm}
    \mathbf{H}_Q=\mathit{Transformer}(\mathit{BERT}(<Q'_i, Q_i>))
\end{align}


\subsection{LLM-based General Evaluation}
The two modules described above mainly concentrate on the content-related similarity assessment, which may fail to assess non-content aspects such as overall presentation-- a factor often considered by human graders, especially for open-ended question requiring structured response.

To address this, the LLM-based general evaluation module is designed to mimic human grading behavior to perform a holistic evaluation by considering not only the correctness or relevance of the student answer but also its presentation quality.
In particular, given the question $Q$, the reference answer $R$ and the student answer $A$, we design prompt to guide the LLM to assess $A$ from multiple perspectives including: {logical soundness}, linguistic fluency and error identification.
The LLM is instructed to generate a concise, holistic assessment that includes: (1) a general evaluation of the answer's overall quality (\eg, very good, good, fair, poor, very poor) and (2) a summary of strengths (\eg, organized presentation) and weaknesses (\eg, logical flaws, inconsistencies). 

The textual assessment generated by the LLM is then tokenized and passed through a BERT model and Transformer encoder to obtain a semantic presentation $\mathbf{H}_G$ for further processing. 

\subsection{Textual Similarity Matching}
While LLM-based evaluation modules excel in capturing deep semantic alignment, they may overlook subtle details due to abstraction and summarization, which can be critical in human grading. Additionally, human graders typically begin by assessing textual similarity between a student's answer and the reference answer as a foundational step before delving into deeper evaluation aspects.

To address this, we integrate a direct textual matching module into our framework alongside the LLM-based modules. This module serves two key purposes: (1) It simulates the initial step of human grading by providing a basic score based on textual similarity, and (2) it complements LLM-based evaluations by capturing important details that might otherwise be missed.

This module is lightweight, employing straightforward text matching techniques: First, a preprocessing step is applied to remove noise and truncate lengthy sentences in the student and reference answers. Next, the cleaned student $A'$ and reference answers $R'$ are concatenated and tokenized:  
\begin{align}
    \mathcal{T} = \mathit{Tokenizer}([CLS], A', [SEP], R')
\end{align}
The tokenized input is then fed into a BERT encoder to generate contextual embeddings, and passed into a Transformer encoder to further model interdependencies between $A'$ and $R'$.
\begin{align}
    \mathbf{H}_T=\mathit{Transformer}(BERT(\mathcal{T}))
\end{align}

Example prompts for each module can be found in the appendix of the extended version~\cite{paper}.
\subsection{Cross-attention Deep Fusion and Prediction}
To unify the insights of the four specialized modules into a holistic grading decision, we introduce a deep fusion and prediction layer that leverages a Transformer encoder to model cross-module dependencies and an MLP layer for final score prediction. 

Specifically, the representation vectors from each module are concatenated along the sequence dimension:
\begin{align}
    \mathbf{H'}= \mathit{Concatenate} (\mathbf{H}_K,\mathbf{H}_Q,\mathbf{H}_G,\mathbf{H}_T)
\end{align}
where $\mathbf{H}\in \mathbb{R}^{4L\times d}$ is stacked outputs from all modules with a sequence length of $4L$.

The concatenated representation is passed into a Transformer encoder that weights and fuses the information across the modules: 
\begin{align}
    \mathbf{H}=\mathit{Transformer}(\mathbf{H'})
\end{align}
The attention mechanism in the Transformer allows the model to capture the cross-module dependencies and attentively integrate the complementary insights from different modules.

Next, the fused embedding $\mathbf{H}$ is compressed into a global representation through mean pooling operation and then passed into a MLP layer for the final score regression:
\begin{align}
    \hat{y}= \mathit{Sigmoid} (\mathit{MLP}(\mathit{MeanPooling}(\mathbf{H})))
\end{align}
where $\hat{y}$ is the predicted score, normalized between 0 and 1 by applying the sigmoid activation.

\sstitle{Loss Function.} Our model, particularly the BERT and Transformer encoders, is trained with the objective of minimizing the difference between the predicted and actual scores for grading tasks. Given a set of $N$ samples with ground-truth scores provided by human graders, we calculate the Mean Squared Error loss to quantify the difference between predicted score $\hat{y}_i$ and actual score $y_i$ as:
\begin{align}
    L=\frac{1}{N}\sum_{i=1}^N (\hat{y}_i - y_i)^2
\end{align}


\section{Experiments}
\label{sec:expt}

\subsection{Experimental settings}
\stitle{Datasets.} To comprehensively evaluate the performance of our proposed framework across different question types and domains, we conduct experiments on two \textit{constructed datasets} covering different types of questions and a \textit{public dataset} focusing on essay-writing questions. The characteristics of datasets is summarized in Table~\ref{tbl:dataset}.
\begin{itemize}[leftmargin=*]
    \item \textbf{General-Type Dataset (GT)}: GT is a constructed dataset featuring a wide variety of subjective questions across four domains. It is generated using LLMs (GPT-4o for knowledge points generation, Qwen-2.5-72B for question generation, and GPT-4o-Mini for answer generation), with manual validation to ensure quality. The dataset includes zero-score, full-score, and partially correct answers for each question. Detailed creation process of the dataset is provided in the appendix of the extended version~\cite{paper}.
    \item \textbf{Domain-Specific Dataset (DS)}: DS is a real-world dataset collected from an e-commerce enterprise, consisting of internal training and certification exams. The dataset includes questions from corporate compliance tests, technical skill certifications, and industry-specific assessments. It features a wide range of subjective question types, such as noun explanations, case studies and analytical questions. 
    \item \textbf{Automated Student Assessment Prize Dataset (ASAP)}~\footnote{https://www.kaggle.com/c/asap-aes/data}: ASAP dataset is a publicly available benchmark for automated essay scoring tasks. 
    This dataset is utilized to assess the framework’s performance on essay-writing questions, particularly those lacking concrete reference answers.
\end{itemize}
All datasets are divided into training, validation and testing sets with a ratio of 10:1:1. 
\begin{table*}[h!]
\centering
\caption{Summary of Datasets.}
\resizebox{\textwidth}{!}{
\begin{tabular}{|c|c|c|c|c|c|c|}
\hline
\textbf{Datasets} & \textbf{Question Types} & \textbf{Domains} & \textbf{Size} & \textbf{Language} & \textbf{Data Sources} & \textbf{Focus} \\ \hline
\textbf{GT} & 
\begin{tabular}[c]{@{}c@{}} 
Mixed \\ (short answer, essay, \etc)
\end{tabular} & 
\begin{tabular}[c]{@{}c@{}} 
Education \\ Architecture \\ Computer science \\ Humanities
\end{tabular} & \begin{tabular}[c]{@{}c@{}} 
12,000 \\ questions \end{tabular} & Chinese &
\begin{tabular}[c]{@{}c@{}} 
LLM-generated \\ with \\ manual review
\end{tabular} & 
\begin{tabular}[c]{@{}c@{}} 
General evaluation \\across domains
\end{tabular} \\ \hline

\textbf{DS} & 
\begin{tabular}[c]{@{}c@{}} 
 Mixed \\ (explanation, case study, \etc)
\end{tabular} & 
\begin{tabular}[c]{@{}c@{}} 
Corporate compliance \\ Technical certification
\end{tabular} & \begin{tabular}[c]{@{}c@{}} 33,724 \\ questions \end{tabular}
 & Mixed (Eng/Chn) &
\begin{tabular}[c]{@{}c@{}} 
Enterprise training \\ tests
\end{tabular} & 
\begin{tabular}[c]{@{}c@{}} 
Domain-specific \\ adaptability
\end{tabular} \\ \hline

\textbf{ASAP} & 
\begin{tabular}[c]{@{}c@{}} 
student-written essays
\end{tabular} & Essays & \begin{tabular}[c]{@{}c@{}} 1,982 essays \\ on 8 topics \end{tabular}
 & English &
\begin{tabular}[c]{@{}c@{}} 
Public benchmark \\ dataset
\end{tabular} & 
\begin{tabular}[c]{@{}c@{}} 
Long-form \\ essay evaluation
\end{tabular} \\ \hline
\end{tabular}
}
\label{tbl:dataset}
\end{table*}


\stitle{Baselines.}
To thoroughly evaluate our proposed framework, we compare it against a wide range of baselines, including two traditional similarity matching methods, nine deep learning methods, and six state-of-the-art LLM-based approaches.
\begin{itemize}[leftmargin=*]
\item \textbf{TF-IDF + SVR}~\cite{sethi2020support}: A classic regression-based model that represents both student and reference answers using TF-IDF features, followed by a Support Vector Regression (SVR) model to predict scores.
\item \textbf{TF-IDF + LightGBM}~\cite{ke2017lightgbm}: Similar to the above, this method replaces SVR with LightGBM, a fast gradient boosting framework, for score prediction based on TF-IDF vectors.
\item \textbf{TextCNN}~\cite{kim-2014-convolutional}: A convolutional neural network trained on top of pre-trained word vectors for sentence-level classification tasks.
\item \textbf{BiLSTM}~\cite{huang2015bidirectional}: A bidirectional LSTM that processes sequential data in both forward and backward directions to create richer representations and predicts scores via cosine similarity.

    \item \textbf{DSSM}~\cite{huang2013learning}: A deep structured semantic model that encodes the reference answer and student answer independently using bert-base-chinese as the backbone. Each answer is encoded into a vector representation, and the cosine similarity between the two vectors is computed for grading. 
    \item \textbf{BERT-base}~\cite{devlin2018bert}: A pre-trained bert-base-chinese model to directly assess the similarity between the student answer and the reference answer by processing them as a text pair. 
    \item \textbf{RoBERTa}~\cite{liu2019roberta}: Similar to BERT, this baseline uses the pre-trained RoBERTa model to process the input text pairs and generates a similarity score between 0 and 1.
    \item \textbf{StructBERT}~\cite{wang2019structbert}: A general-purpose model for text similarity tasks. This baseline converts continuous scores into binary labels (0 or 1) and trains the model as a binary classifier to predict if the answers belong to the same class.
    \item \textbf{BGE-base, BGE-reranker}~\cite{xiao2024c}: These models use the BGE-base-zh-v1.5 and BGE-reranker to assess the similarity between student and reference answers.

\item \textbf{NPCR}~\cite{xie2022automated}: A neural pairwise contrastive regression that combines regression and ranking objectives through contrastive learning to enhance scoring stability and accuracy.

    \item \textbf{Qwen2.5 series}~\cite{yang2024qwen2}: Qwen2.5-72B-Instruct and Qwen2.5-14B-Instruct models are used to generate a score based on the input, which includes the question, student answer, and reference answer. In the Qwen2.5-72B-Instruct-COT variant, the model is enhanced with a Chain-of-Thought (COT) reasoning prompt, instructing the LLM to reason step-by-step before generating a score. 
    \item \textbf{ChatGLM3-6B}~\cite{zeng2022glm}: A 6-billion parameter bilingual LLM developed by Tsinghua University’s KEG Lab and Zhipu AI bilingual. 

\item \textbf{Baichuan2-13B-chat-v1}~\cite{baichuan2023open}: A 13-billion parameter instruction-tuned LLM developed by Baichuan Inc. for Chinese-English bilingual tasks.

\item \textbf{DeepSeek-v3}~\cite{deepseek2024}: A recent multilingual instruction-tuned model developed by DeepSeek AI. 
\end{itemize}

\stitle{Implementation.} 
For data processing, we utilize the BERT Tokenizer to encode the text, with a maximum sequence length truncated to 128 tokens. During data loading, we use the DataLoader module to set a batch size of 20 for training and 64 for validation. In each LLM-based module, the Qwen2.5-72B-Instruct model is employed to generate key points, general evaluation and pseudo-questions.
For model training, we use a learning rate of $2e-5$ and the AdamW optimizer. The training consists of 10 epochs, and the batch size is set to 20 for the training dataset and 64 for the validation dataset. The parameters of baselines are set as suggested in the original papers.
 
\stitle{Metrics.}
We employ a comprehensive set of metrics to evaluate the performance of our framework on handling various grading scenarios, from regression (MSE) to binary (ACC and F1) and ordinal classification evaluations (QWK).

First, we adopt Mean Squared Error (MSE) to measure the difference between the predicted score $\hat{y}$ and the ground-truth score $y$: 
\vspace{-3mm}
\begin{align}
\vspace{-3mm}
    \text{MSE}=\frac{1}{N}\sum_{i=1}^N(\hat{y}_i-y_i)^2
\vspace{-3mm}
\end{align}
\vspace{-3mm}

For binary scenarios (\eg, pass/fail), we convert scores to binary labels ($\hat{y}_i \geq 0.5$ is correct) and report the F1 score, the harmonic mean of precision and recall, and Accuracy (ACC):
\vspace{-2mm}
\begin{align}
    \text{ACC}=\frac{1}{N}=\sum_{i=1}^N \mathbb{I}(round(\hat{y}_i)=y_i)
    \vspace{-3mm}
\end{align}
where $\mathbb{I}$ is an indicator function returning 1 if the condition is true and 0 otherwise.

Finally, to measure agreement with human graders in an ordinal setting, we use Quadratic Weighted Kappa (QWK)~\cite{xie2022automated}. We discretize scores into five levels $[0.0, 0.25, 0.5, 0.75, 1.0]$:
\vspace{-2mm}
\begin{align}
    \text{QWK} = 1-\frac{\sum_{i,j}\mathbf{w}_{ij}\mathbf{o}_{ij}}{\sum_{i,j}\mathbf{w}_{ij}\mathbf{e}_{ij}}
    \vspace{-3mm}
\end{align}
where $\mathbf{o}_{ij}$ is the observe agreement (the number of cases graded as $i$ by human and $j$ by the automated system), $\mathbf{e}_{ij}$ is the expected agreement assuming random assignment, and $\mathbf{w}_{ij}$ represents quadratic weights penalizing large grading deviations. QWK ranges from -1 (worse than random) to 1 (perfect agreement), with higher value indicating better human-algorithm consistency.

\subsection{Sensitivity of hyper-parameters}
\begin{figure}[h]
\centering
    \includegraphics[width=\linewidth]{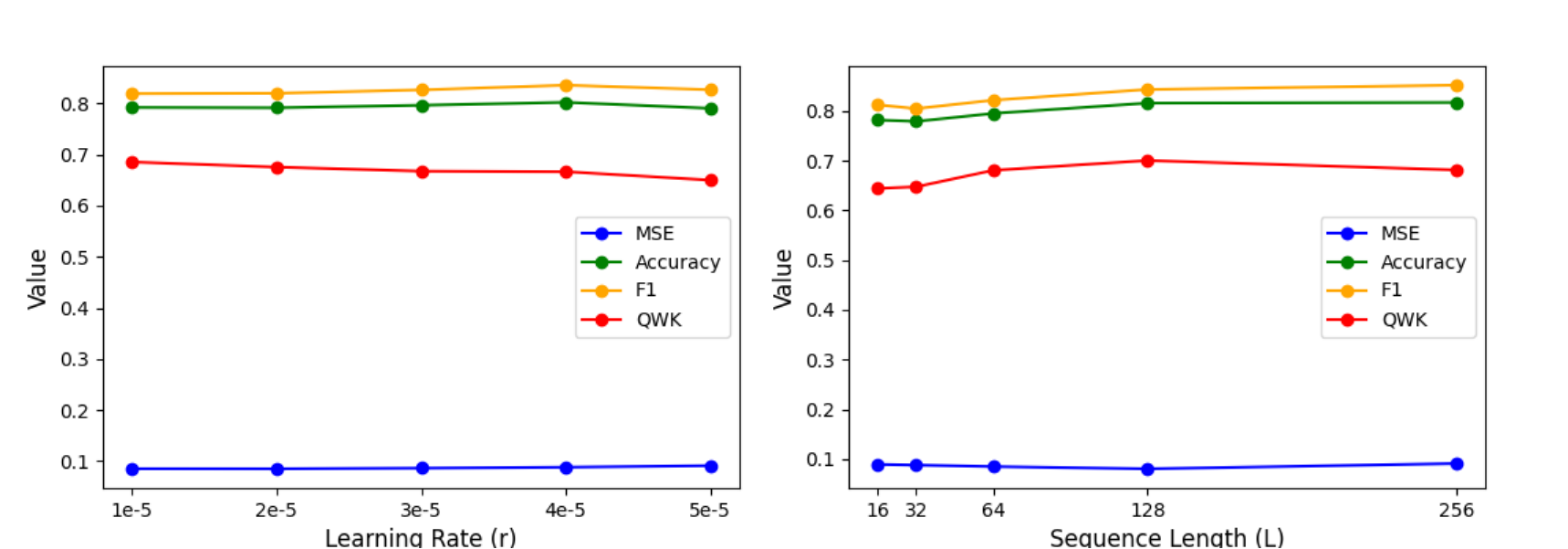}
\caption{Impact of Hyper-parameters.}
\vspace{6mm}
\label{fig:para}
\end{figure}

We analyze the impact of hyper-parameters, including the learning rate and the sequence length $L$, on the performance of our framework. Due to space limitation, we present the results only for the DS dataset, though similar trends are observed across other datasets. As shown in Fig.~\ref{fig:para}, our method performs optimally when the learning rate $r=2e-5$ and sequence length $L=128$. These settings are adopted for the subsequent experiments.

\subsection{Performance Analysis}
We compare our method against a wide range of baselines, and summarize the results in Table~\ref{tab:comparison}, where the best results are bolded and the second-best results are underlined. 
Our observations and analyses are as follows:
(1) \emph{Analysis across metrics:} Our framework consistently achieves superior performance across all metrics, particularly excelling in stricter metrics MSE and QWK. For example, on the DS dataset, which features the most complex technical exam data, our model significantly outperforms the strongest learning-based baseline (NPCR) and LLM-based baseline (Qwen2.5-72B), improving the QWK by 8.5\% and 31.9\% respectively. 
(2) \emph{Analysis across datasets:} Our model demonstrates strong and consistent performance across all datasets. On the GT dataset, where questions and answers are LLM-generated and well-structured, all models perform well- particularly LLM-based ones. Yet, our model achieves a notable MSE drop over the strongest LLM-based baseline (0.019 vs. 0.038).
In the DS dataset, which poses the greatest challenge due to varied formats, mismatched answer lengths, and multi-lingual content, most models perform worse. Our model still leads, decreasing MSE to 0.080 and improving ACC, F1, QWK to 0.816, 0.843 and 0.700 respectively, highlighting its robustness in real-world assessment tasks.
(3) \emph{Analysis across LLM complexity:} Our approach maintains robust performance even with lighter LLMs (\eg, Qwen2.5-7B), surpassing traditional and learning-based baselines across all datasets. This highlights the performance gains primarily stem from our module design and deep fusion strategy rather than from model scale alone. 

\stitle{Latency \& cost.} 
Our model is deployed on Alibaba Cloud with an API QPS (Questions Per Second) of 20, allowing parallel execution of all LLM modules. The total latency for grading a single question is approximately 3-5 seconds: \~0.2 seconds for the BERT and Transformer layers and 3-4 seconds for the most expensive LLM module. This latency is acceptable in practical use, especially since the system is designed for asynchronous batch grading, where grading is performed in the backend and users receive the results once processing is complete.

\begin{table*}[htbp]
\centering
\caption{Comparison of Model Performance on GT, DS, and ASAP Datasets}
\label{tab:comparison}
\resizebox{\textwidth}{!}{
\begin{tabular}{|c|l|cccc|cccc|cccc|}
\hline
  \multirow{2}{*}{\textbf{Category}} & \multirow{2}{*}{\textbf{Model}} & \multicolumn{4}{c|}{\textbf{GT Dataset}} & \multicolumn{4}{c|}{\textbf{DS Dataset}} & \multicolumn{4}{c|}{\textbf{ASAP Dataset}} \\
\cline{3-14}
 &  & MSE$\downarrow$ & ACC$\uparrow$ & F1$\uparrow$ & QWK$\uparrow$ & MSE$\downarrow$ & ACC$\uparrow$ & F1$\uparrow$ & QWK$\uparrow$ & MSE$\downarrow$ & ACC$\uparrow$ & F1$\uparrow$ & QWK$\uparrow$ \\
\hline
\multirow{2}{*}{Traditional} 
 & TF-IDF + SVR & 0.102 & 0.750 & 0.761 & 0.562 & 0.095 & 0.785 & 0.821 & 0.613 & 0.069 & 0.766 & 0.748 & 0.654 \\
 & TF-IDF + LightGBM & 0.088 & 0.756 & 0.791 & 0.647 & 0.097 & 0.784 & 0.824 & 0.608 & 0.066 & 0.759 & 0.742 & 0.695 \\
\hline
\multirow{9}{*}{Learning-based} 
 & TextCNN & 0.088 & 0.764 & 0.787 & 0.671 & 0.100 & 0.775 & 0.813 & 0.615 & 0.074 & 0.744 & 0.722 & 0.661 \\
 & BiLSTM & 0.087 & 0.803 & 0.820 & 0.700 & 0.144 & 0.689 & 0.785 & 0.329 & 0.087 & 0.693 & 0.651 & 0.596 \\
 & BERT-Base & 0.032 & 0.878 & 0.892 & 0.752 & 0.090 & 0.797 & 0.827 & 0.583 & {0.060} & 0.762 & 0.738 & 0.533 \\
 & RoBERTa & 0.038 & 0.873 & 0.893 & 0.738 & {0.087} & {0.802} & {0.833} & 0.591 & 0.066 & 0.763 & 0.739 & 0.535 \\
 & StructBERT & 0.126 & 0.776 & 0.721 & 0.613 & 0.192 & 0.646 & 0.561 & 0.328 & 0.189 & 0.755 & 0.308 & 0.196 \\
 & BGE-base & 0.197 & 0.612 & 0.739 & 0.285 & 0.176 & 0.640 & 0.746 & 0.238 & 0.174 & 0.648 & 0.741 & 0.268 \\
 & BGE-reranker & 0.286 & 0.672 & 0.769 & 0.267 & 0.219 & 0.694 & 0.759 & 0.372 & 0.244 & 0.681 & 0.741 & 0.374 \\
 & DSSM & 0.073 & 0.805 & 0.836 & 0.728 & 0.104 & 0.767 & 0.809 & {0.611} & 0.062 & 0.780 & 0.770 & {0.734} \\
 & NPCR & 0.055 & 0.823 & 0.842 & 0.667 & 0.095 & 0.774 & 0.805 & {0.615} & \underline{0.060} & 0.792 & 0.782 & {0.749} \\
\hline
\multirow{6}{*}{LLM-based}
 & Qwen2.5-14B & 0.040 & \underline{0.914} & {0.925} & 0.876 & 0.156 & 0.684 & {0.762} & 0.364 & 0.103 & 0.799 & {0.819} & 0.588 \\
 & Qwen2.5-72B & 0.038 & \textbf{0.917} & \underline{0.927} & {0.881} & 0.156 & 0.685 & 0.759 & 0.381 & 0.102 & 0.792 & {0.817} & 0.588 \\
 & Qwen2.5-72B-COT & 0.041 & \textbf{0.917} & \underline{0.927} & 0.866 & 0.156 & 0.681 & 0.759 & 0.361 & 0.096 & {0.799} & {0.817} & 0.615 \\
 & Chatglm3-6B & 0.040 & 0.908 & 0.920 & 0.872 & 0.157 & 0.680 & 0.759 & 0.362 & 0.101 & {0.801} & \underline{0.821} & 0.593 \\
 & Baichuan2-13B-chat-v1 & 0.040 & 0.910 & 0.922 & 0.875 & 0.158 & 0.680 & 0.758 & 0.359 & 0.101 & {0.801} & \underline{0.821} & 0.593 \\
 & Deepseek-v3 & 0.040 & 0.907 & 0.919 & 0.872 & 0.156 & 0.682 & 0.760 & 0.362 & 0.104 & 0.796 & 0.816 & 0.581 \\
\hline
\multirow{3}{*}{LLM-enhanced} & Ours-Qwen2.5-7B 
&0.024 &0.903&0.913&0.912
& 0.088 & \underline{0.809}& \textbf{0.845}& 0.676
& \underline{0.060} & 0.816 & 0.816 &0.750\\ 

& Ours-Qwen2.5-14B 
& \underline{0.021} & 0.898 & 0.907 &\underline{0.915}
&\underline{0.086} & 0.802 & 0.832 & \underline{0.681}
& \underline{0.060} & \underline{0.817} & 0.816 & \underline{0.760}\\
& Ours-Qwen2.5-72B & \textbf{0.019} & \textbf{0.917} & \textbf{0.928} & \textbf{0.929} & \textbf{0.080} & \textbf{0.816} & \underline{0.843} & \textbf{0.700} & \textbf{0.059} & \textbf{0.836} & \textbf{0.840} & \textbf{0.772} \\
\hline
\end{tabular}
}
\end{table*}

\begin{table*}[htbp]
\centering
\caption{Automated Grading Examples}
\label{tab:grading_examples}
\resizebox{\textwidth}{!}{
\begin{tabular}{|l|p{5.5cm}|p{6.5cm}|p{5.6cm}|}
\hline
\textbf{} & \begin{tabular}[c]{@{}c@{}} {\textbf{Case 1: High-Scoring Example}} \end{tabular} & \textbf{Case 2: Medium-Scoring Example} & \textbf{Case 3: Low-Scoring Example} \\
\hline
\textbf{Question} & Explain the main differences between PPP and BOT models. & Describe the basic components and functions of a LAN. & Explain the advantages and applications of PPP in construction projects. \\
\hline
\textbf{Student Answer} & PPP emphasizes long-term collaboration with government while BOT focuses on phased transfer before transferring it to the government. & LAN consists of devices/adapters for data exchange. Its main functions are data exchange and resource sharing. & PPP improves efficiency in commercial sectors. \\
\hline
\textbf{Reference Answer} & PPP: long-term partnership; BOT: specific-phase operation. & LAN includes servers, workstations, and protocol; enables resource sharing, communication, and service sharing. & PPP advantages: risk-sharing, higher service quality, diversified funding (reduces government burden). \\
\hline
\textbf{Key Points (Student)} & PPP: long-term; BOT: transfer & Devices, adapters; data exchange & Efficiency \\
\hline
\textbf{Key Points (Reference)} & PPP: partnership; BOT: phased operation & Servers, workstations; communication; sharing & Risk-sharing; service quality; funding \\
\hline
\textbf{Pseudo-Question} & Explain differences between PPP and BOT, focusing on cooperation methods and implementation. & List LAN components and describe their primary functions. & Describe PPP's main roles and application fields in projects. \\
\hline
\textbf{LLM Evaluation} & Good: Correctly identifies core differences but lacks depth on PPP collaboration scope. & Average: Captures basic elements but misses some components like servers/protocols and functionalities. & Poor: Misses critical points (risk-sharing, funding) and misrepresents applications. \\
\hline
\textbf{Predicted / True Scores} & 0.99 / 1.0 & 0.52 / 0.5 & 0.01 / 0.0 \\
\hline
\end{tabular}
}
\end{table*}

\subsection{Ablation Study}
We conduct an ablation study on GT dataset to investigate the contribution of each key module. 
We compare our full model with five variants: (1) removing key points matching module (w/o KPM); (2) removing pseudo-question matching module (w/o PQM); (3) removing LLM general evaluation (w/o LGE); (4) removing text similarity matching module (w/o TSM); (5) replacing cross-module deep fusion with simple feature concatenation from different modules (w/o Cross).

Based on the ablation results in Table~\ref{tbl:ablation}, we draw the following observations: (1) The full model outperforms all ablated variants, confirming that each module contributes unique perspectives on grading, and that the cross-module fusion is essential for synthesizing them into an accurate, human-like score.
(2) Removing the cross-module fusion leads to the most significant performance drop across all metrics, indicating that transformer-based fusion is crucial for effectively integrating signals from all modules and producing coherent grading signals.
(3) Excluding either LGE or KPM module leads to consistent performance drops. This highlights their importance in modeling semantic correctness and grading criteria at a conceptual level- key aspects of human-like evaluation.
(4) The PQM and TSM modules show minor impact when removed. This is understandable in the GT dataset, where both questions and answers are LLM-generated and thus typically well-aligned. In such cases, reverse matching and text-level similarity offer limited additional gain. Nonetheless, their presence still contributes to the overall performance.


\begin{table}[htbp]
\centering
  \caption{Results for ablation study.}
  \label{tbl:ablation}
  \resizebox{0.8\linewidth}{!}{
  \begin{tabular}{|c|c|c|c|c|}
    \hline
    Model & \textbf{MSE} $\downarrow$ & \textbf{ACC} $\uparrow$ & \textbf{F1} $\uparrow$ & \textbf{QWK} $\uparrow$\\
    \hline
    w/o LGE & 0.022 & 0.911 & 0.920 & 0.920 \\
    w/o PQM & 0.019 & 0.916 & 0.927 & 0.926 \\
    w/o KPM & 0.020 & 0.907 & 0.916 & 0.923 \\
    w/o TSM & 0.019 & 0.916 & 0.926 & 0.926 \\
    w/o Cross & 0.020 & 0.894 & 0.905 & 0.915 \\
    \hline
    Full & \textbf{0.018} & \textbf{0.917} & \textbf{0.928} & \textbf{0.929} \\
    \hline
  \end{tabular}
  }
\end{table}

\subsection{Case study}
We further conduct a qualitative analysis through representative examples to demonstrate how our model effectively simulates human grading.
 Table~\ref{tab:grading_examples} shows the three examples representing high, medium and low scoring scenarios from GT dataset while more examples on AES can be found in the appendix of the extended version~\cite{paper}.
 
In Case 1 (High-Scoring), the student's response effectively captured the core distinction between PPP and BOT models. Our model precisely identified these valid key points, assigning a high predicted score (0.99), closely aligning with human grading (1.0). In the medium-scoring case, the student partially covered basic LAN components (devices/adapters) and functions (data exchange, resource sharing) but omitted critical aspects like servers and communication protocols. Our model accurately identified these omissions, assigning a moderate predicted score (0.52), well-matched to the human score (0.5). In Case 3 (Low-scoring), the response provided a superficial answer about PPP efficiency without addressing crucial points like risk-sharing and service quality. The model identified these severe inadequacies, assigning a very low score (0.01) consistent with human grading (0.0).

Overall, our auto-grading approach demonstrates consistent and accurate performance in identifying critical elements across varied response quality, closely matching human evaluation standards.

\subsection{Online A/B Test}
After a comprehensive offline evaluation, the proposed auto-grading framework has been successfully deployed in Alibaba's online testing system to grade real-world tests. To assess its performance in practical scenarios, we compare it against a similarity matching-based grading baseline that utilizes BERT model for scoring.

We conduct experiments on two distinct categories of tests:
\vspace{-2mm}
\begin{itemize}[leftmargin=*]
    \item \textbf{Value-oriented tests (D1)}: This dataset includes 100 subjective questions focused on personal reflection and value-oriented assessments. These questions are designed to evaluate candidates' compliance knowledge, decision-making skills, and alignment with personal and organizational values.
    \item \textbf{Technique-oriented tests (D2)}: This dataset consists of 2,000 questions aimed at assessing knowledge, problem-solving, and technical proficiency in areas such as operations, customer service, finance, and other enterprise-related tasks. The questions cover a variety of practical and scenario-based challenges, requiring multi-step reasoning and decision-making.
\end{itemize}
 
\begin{table}[htbp]
\centering
\caption{Online Evaluation of Model Performance}
\label{tbl:online}
 \resizebox{\linewidth}{!}{
\begin{tabular}{|l|cccc|cccc|}
\hline
 \multirow{2}{*}{\textbf{Model}} & \multicolumn{4}{c|}{\textbf{D1 Dataset}} & \multicolumn{4}{c|}{\textbf{D2 Dataset}} \\
 \cline{2-9}
 & MSE$\downarrow$ & ACC$\uparrow$ & F1 $\uparrow$ & QWK$\uparrow$ & MSE$\downarrow$ & ACC$\uparrow$ & F1 $\uparrow$ & QWK$\uparrow$ \\
\hline
Baseline & 0.130 & 0.760 & 0.807 & 0.597 
& 0.079 & 0.794 & 0.829 & 0.691 \\
Ours-Qwen2.5-72B & \textbf{0.125} & \textbf{0.780} & \textbf{0.823} & \textbf{0.610} 
& \textbf{0.078} & \textbf{0.814} & \textbf{0.844} & \textbf{0.704} \\
\hline
\end{tabular}
}
\end{table}
From the results illustrated in Table ~\ref{tbl:online}, we can conclude that our method is consistently superior on all type of grading tasks in real online testing scenarios.

\section{Conclusion}
In this paper, we present a unified LLM-enhanced auto-grading framework that addresses the challenges of grading diverse types of subjective questions. By integrating four complementary modules, our framework provides a holistic, human-like evaluation of student answers from multiple dimensions. Experimental results on general and domain-specific datasets demonstrate that our method significantly outperforms baseline approaches across various question types. The successful deployment of our solution in real-world online testing platform within a leading e-commerce enterprise highlights its practical effectiveness and robustness.

\section*{Acknowledgments}
This work is partially supported by the Zhejiang Provincial Natural Science Foundation (No. LY24F020013) and Alibaba Innovative
Research Program. The authors would like to acknowledge the
Supercomputing Center of Hangzhou City University and Zhejiang
Provincial Engineering Research Center for Real-Time SmartTech in Urban Security Governance, for their support of the advanced
computing resources.
\bibliography{sample-base.bib}

\end{document}